\documentclass[10pt,twocolumn,letterpaper]{article}

\usepackage[review]{cvpr}      

\definecolor{cvprblue}{rgb}{0.21,0.49,0.74}
\usepackage[pagebackref,breaklinks,colorlinks,allcolors=cvprblue]{hyperref}

\usepackage{amsmath}
\usepackage{amssymb}
\usepackage{multirow}
\usepackage{booktabs}
\usepackage{colortbl}
\usepackage{xcolor}
\usepackage{array}
\usepackage{pifont}
\usepackage{bm}
\usepackage[misc]{ifsym}

\def\eg{{\it{e.g.}}}

\def\ie{{\it{i.e.}}}
\definecolor{mycolor}{RGB}{241,240,255}

\def\paperID{10018} 
\def\confName{CVPR}
\def\confYear{2025}

\title{PMA: Towards Parameter-Efficient Point Cloud Understanding via \\ Point Mamba Adapter}

\author{%
    Yaohua Zha$^{1,2}$
  ~~ Yanzi Wang$^1$ 
  ~~ Hang Guo$^1$
  ~~ Jinpeng Wang$^1$
  ~~ Tao Dai$^{3,}\thanks{Corresponding author. ${\text{\Letter}}$ daitao.edu@gmail.com}$  \\
  ~~ Bin Chen$^4$
  ~~ Zhihao Ouyang$^5$ 
  ~~ Xue Yuerong$^1$ 
  ~~ Ke Chen$^2$
  ~~ Shu-Tao Xia$^{1,2}$ \\
  \normalsize $^1$Tsinghua University 
  ~~ $^2$Pengcheng Laboratory
  ~~ $^3$Shenzhen University \\
  \normalsize ~~ $^4$Harbin Institute of Technology, Shenzhen
  ~~ $^5$Meta\\
}

\begin{document}
\maketitle
\begin{abstract}
Applying pre-trained models to assist point cloud understanding has recently become a mainstream paradigm in 3D perception. However, existing application strategies are straightforward, utilizing only the final output of the pre-trained model for various task heads. It neglects the rich complementary information in the intermediate layer, thereby failing to fully unlock the potential of pre-trained models. To overcome this limitation, we propose an orthogonal solution: \textbf{P}oint \textbf{M}amba \textbf{A}dapter (\textbf{PMA}), which constructs an ordered feature sequence from all layers of the pre-trained model and leverages Mamba to fuse all complementary semantics, thereby promoting comprehensive point cloud understanding.
Constructing this ordered sequence is non-trivial due to the inherent isotropy of 3D space. Therefore, we further propose a geometry-constrained gate prompt generator (G2PG) shared across different layers, which applies shared geometric constraints to the output gates of the Mamba and dynamically optimizes the spatial order, thus enabling more effective integration of multi-layer information.
Extensive experiments conducted on challenging point cloud datasets across various tasks demonstrate that our PMA elevates the capability for point cloud understanding to a new level by fusing diverse complementary intermediate features. Code is available at \url{https://github.com/zyh16143998882/PMA}. 
\end{abstract}    
\section{Introduction}

Point clouds, as a fundamental 3D representation, have been widely applied in fields such as autonomous driving and robotics, drawing considerable attention. These successful applications have largely benefited from advancements in deep learning-based point cloud understanding~\cite{pointnet,dgcnn,ptv2,ptv3,pointmamba,lcm}. Existing deep learning-based approaches for point cloud can be broadly categorized into supervised end-to-end paradigms~\cite{pointnet++,3detr,pointmlp,sfr} and self-supervised pre-training \& fine-tuning paradigms~\cite{pointbert,pointmae,femae,recon,pointgpt}. The latter, by fine-tuning general representations from pre-trained point cloud models, has significantly advanced the capability for point cloud understanding and has become the mainstream paradigm in the field.

To leverage pre-trained models, most existing works~\cite{pointbert,pointmae,recon,pointgpt} adopt a Fully Fine-Tuning (FFT) strategy, where the parameters of both the pre-trained model and task head are updated simultaneously during downstream tasks, yielding satisfactory performance and convergence speed. However, this approach is resource-intensive, as it requires learning unique parameters for each task and dataset, resulting in significant memory and storage demands. Some efforts \cite{idpt, ppeft, dapt} have introduced Parameter-Efficient Fine-Tuning (PEFT) to point cloud models, mitigating resource demands by freezing the weights of pre-trained models and tuning only the task heads and additional learnable parameters.

Although these works achieve a balance between efficiency and performance, they still fail to deliver satisfactory performance across various tasks, especially in fine-grained point cloud understanding tasks such as segmentation. 
This is because segmentation requires fine-grained understanding at the point level, whereas classification involves a global, coarse-grained understanding of the point cloud.
Existing PEFT methods rely on adapting various tasks with only a small number of additional learnable parameters. This approach heavily depends on the final output of the frozen pre-trained model. However, as shown in Figure \ref{obs}, our experimental observations demonstrate that the intermediate features of the pre-trained model contain information that is almost comparable to that of the final features. Therefore, the existing PEFT practice of \textit{completely discarding intermediate features and relying solely on the final features fails to fully exploit the understanding capabilities of the pre-trained model}.

To overcome this limitation, we propose an orthogonal PEFT solution based on the Mamba architecture: \textbf{P}oint \textbf{M}amba \textbf{A}dapter (\textbf{PMA}), which sequentially integrates all intermediate layer features for downstream tasks, thereby unlocking the potential of the pre-trained model.
The core innovation of PMA lies in constructing an ordered sequence of features extracted from all layers of the pre-trained backbone and leveraging state space models to fuse complementary semantics.
However, constructing this ordered sequence is non-trivial due to the inherent isotropy of 3D space. We introduce a geometry-constrained gate prompt generator (G2PG) shared across different layers, which applies shared geometric constraints to the output gates of the Mamba. This approach dynamically optimizes spatial ordering and improves the adaptability of the Mamba’s out, allowing us to adaptively adjust the sequence for different tasks in an end-to-end manner, thus enabling more effective integration of multi-layer information.
Extensive experiments on challenging point cloud datasets across various tasks demonstrate that our PMA significantly enhances point cloud understanding by efficiently fusing diverse intermediate features, leading to state-of-the-art performance in multiple tasks.

The main contributions can be summarized as follows:
\begin{itemize}
    \item  We propose the Point Mamba Adapter (PMA), an innovative parameter-efficient fine-tuning framework that effectively integrates intermediate features from pre-trained models to improve 3D point cloud understanding, enabling efficient and comprehensive feature fusion.
    \item To address the inherent isotropy in 3D space, we introduce a geometry-constrained gate prompt generator (G2PG) shared across different layers, which applies shared geometric constraints to the output gates of the Mamba and dynamically optimizes the spatial order, thus enabling more effective integration of multi-layer information.
    \item Extensive experiments of downstream tasks demonstrate that PMA achieves significant improvements in understanding 3D point clouds, particularly for tasks requiring fine-grained, point-level understanding.
\end{itemize}

\section{Related Work}

\subsection{Parameter-Efficient Transfer Learning}

As pre-trained models grow larger, Parameter-Efficient Transfer Learning (PETL), which aims at reducing memory and storage costs by updating only a fraction of model parameters, has become essential across NLP~\cite{adaptformer,ding2023parameter,promptuning,adapter,Delta-tuning}, 2D computer vision~\cite{vpt,ptuningv2,coop,cocoop,Tip-Adapter,zhang2024parameter,adapterir}, and increasingly, 3D vision~\cite{p2p,idpt,dapt,ppeft}. PEFT methods mainly fall into Prompt-based~\cite{promptuning,vpt,ptuningv2,tsimpoukelli2021multimodal,idpt} and Adapter-based~\cite{adapter,adaptformer,adapter,adapterir} categories. Prompt-based approaches add learnable tokens to input data or attention layers to guide task-specific tuning. For example, Visual Prompt Tuning (VPT~\cite{vpt}) enhances ViTs by adding tunable tokens to input or hidden layers. In contrast, Adapter-based methods introduce lightweight modules, such as adapters within frozen model layers. Techniques like AdaptFormer~\cite{adaptformer} add adapters in each Feed-Forward Network (FFN) layer, and LoRA~\cite{lora} employs low-rank approximations to weight matrices, reducing parameters while maintaining performance. 

In the 3D field, some efforts~\cite{idpt,ppeft,dapt,sun2024parameter,pointgst} have introduced Parameter-Efficient Fine-Tuning (PEFT) to point cloud. IDPT~\cite{idpt} first attempt introduces an instance-aware dynamic prompt tuning to pre-trained point cloud models, Point-PEFT~\cite{ppeft} and DAPT~\cite{dapt} simultaneously apply prompt tuning combined with additional Adapters to point clouds, significantly reducing additional parameters and delivering notable performance gains. Although these works achieve a balance between efficiency and performance, they still fail to deliver satisfactory performance across various tasks, especially in fine-grained point cloud understanding tasks. We aim to fully unleash the potential of pre-trained models by designing a PEFT method capable of efficiently and comprehensively integrating features across all layers of the pre-trained model.

\subsection{State Space Model}

State Space Models (SSMs) have emerged as a powerful framework for sequence modeling, drawing from control theory to enable efficient and scalable representations~\cite{hippo,l4d,s4,lssl,s5}. By integrating state-based dynamics into neural architectures, SSMs combine the fast inference advantages of RNNs with the parallel training strengths of CNNs, making them ideal for long-sequence processing with linear time complexity. Early developments like the Structured State-Space Sequence model (S4)~\cite{s4} demonstrated the capacity of deep SSMs to capture long-range dependencies effectively. S5~\cite{s5} further optimized SSM architectures by introducing multi-input multi-output (MIMO) configurations and refined parallel processing, enabling even more efficient computation. Recently, Mamba~\cite{mamba}, designed with hardware efficiency and selective state-space processing, has shown performance and efficiency gains over Transformers~\cite{attention}. Its application in the visual domain through adaptations like Vision Mamba~\cite{visionmamba}, VMamba~\cite{vmamba} and MambaIR~\cite{mambair}. PointMamba~\cite{pointmamba}, Mamba3D~\cite{mamba3d}, and LCM~\cite{lcm} extends Mamba to point cloud analysis, where it traverses input sequences along well-designed space direction to effectively capture spatial dependencies in 3D data. In this paper, we introduce Mamba, orthogonal to the pre-trained model, as an Adapter to efficiently fuse features across all layers, thereby fully unleashing the potential of the pre-trained model.







\section{Preliminaries}

State Space Models~\cite{s4} have emerged as a powerful framework for sequence modeling. Its essence lies in representing a continuous system that maps an input sequence $x_t$ to an output sequence $y_t$ through a hidden latent state $h_t \in \mathbb{R}^{S}$, where $S$ is the dimension of the latent state. This sequence-to-sequence mapping can be expressed by the following formula:
\begin{gather}
    \label{eq1}
    h_t = \overline{\bm A}h_{t-1} + \overline{\bm B}x_t \\
    y_t = {\bm C}h_{t} + {\bm D}x_{t} \\
    \overline{\bm A}=exp({\bm A}{\bm \Delta}) \\
    \overline{\bm B}=({\bm A}{\bm \Delta})^{-1}(exp({\bm A}{\bm \Delta})-{\bm I}){\bm \Delta}{\bm B}
\end{gather}
where the $\overline{\bm A} \in \mathbb{R}^{S \times S}$ is the state transition matrix, $\bm B \in \mathbb{R}^{1 \times N}$ is the input-to-state matrix, \eg the the input matrix, and $\bm C \in \mathbb{R}^{1 \times N}$ is the state-to-output matrix, \eg the output matrix. $\bm D \in \mathbb{R}^{1 \times N}$ is a residual connection. 

However, in state space models, the above parameters $(\overline{\bm A},\overline{\bm B},{\bm C},{\bm \Delta})$ are fixed, and this static nature limits their ability to adapt to content-aware modeling. Mamba~\cite{mamba} introduces selectivity by allowing the matrices ${\bm A},{\bm B},{\bm C}$ of the SSM to depend on the input data. This enables the model to dynamically adjust its state based on the current input, selectively propagating or ignoring information as needed.

\section{Methodology}
\subsection{Observation of Intermediate Features}

Previous point cloud PEFT methods~\cite{idpt,ppeft,dapt} typically introduce a small number of additional learnable parameters in each frozen layer of the pre-trained model for layer-wise fine-tuning, then pass the features from the final layer to downstream tasks while discarding the features from all intermediate layers. However, we argue that the discarded intermediate layer features may contain information as semantically important as that of the final layer, such as complementary information. This information is crucial for comprehensive point cloud analysis, especially for a fine-grained understanding of point clouds.

To illustrate our idea, we first design a simple empirical observation experiment to analyze the importance of features from each layer individually. Specifically, we observe the performance on downstream tasks when \textit{not using the full pre-trained model}, but instead using only the first \textit{1} layer, the first \textit{2} layers, and so on, up to all layers of the pre-trained model. We freeze the entire pre-trained model and sequentially pass the embeddings of point patches through the first \textit{n} layers, feeding their output to a downstream task head. During fine-tuning, only the parameters of the task head are updated. We use the pre-trained Point-MAE~\cite{pointmae} model as the backbone and validate our idea on the classification task with the ScanObjectNN~\cite{scanobjectnn} dataset and the part segmentation task with the ShapeNetPart~\cite{shapenet} dataset.

\begin{figure}[t]
\centering
\includegraphics[width=1.0\linewidth]{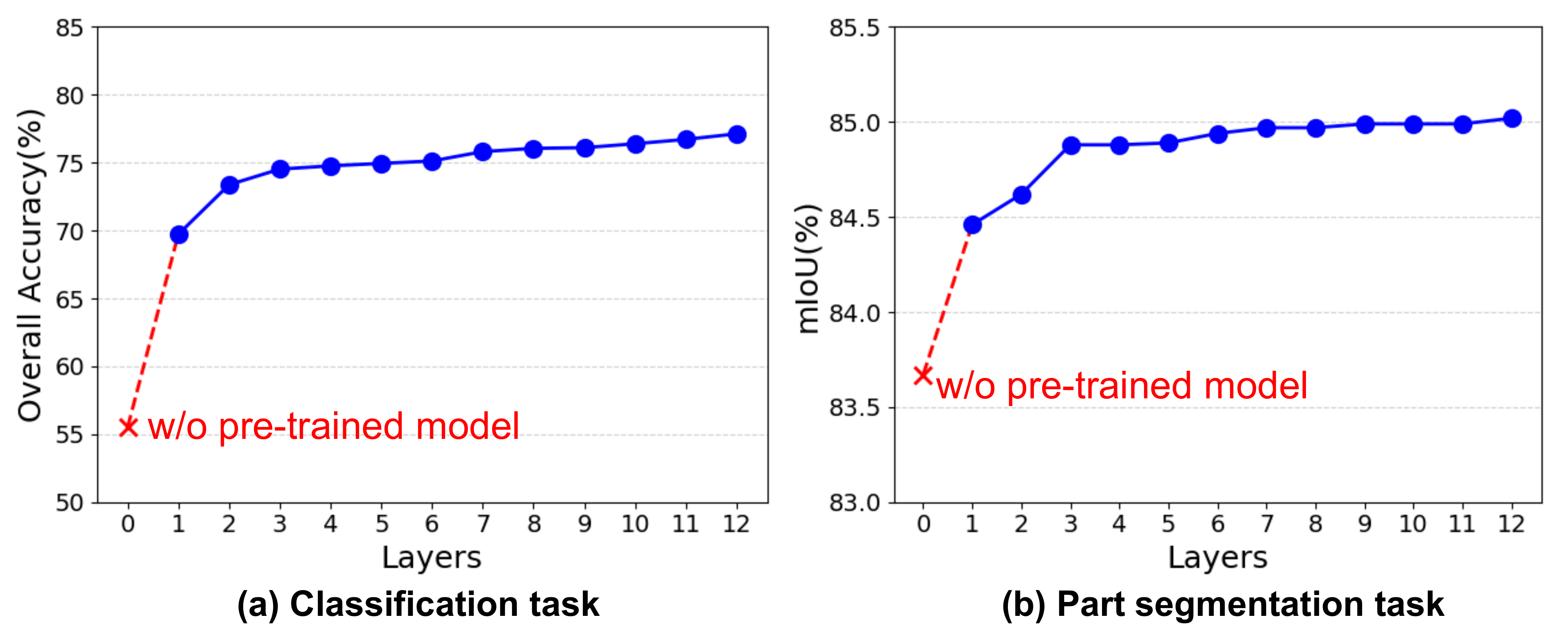} 
\caption{The impact of freezing different numbers of pre-trained model layers on the output features and their effect on downstream tasks. The performance is evaluated using the Point-MAE pre-trained model on classification and part segmentation tasks.
}
\label{obs}
\end{figure}

Figure \ref{obs} shows our observation results, and it is clear that performance steadily improves as the number of layers increases. However, the most notable finding is that \textit{the performance when using only the first few layers is not significantly lower than when using all layers}. For instance, using only the first 3 layers results in a decrease of just 2.6\% and 0.14\% in classification and segmentation tasks, compared to using all layers. Similarly, using the first 6 layers reduces performance by only 2.0\% and 0.09\%. These drops are negligible compared to the significant drop observed when no pre-trained model is used at all (21.63\% and 1.35\%).
From these observations, we argue that the \textbf{\textit{intermediate features of the pre-trained model contain information nearly equivalent to, and even complementary to, the final features}}. It encourages us to design a method that can efficiently and effectively integrate features from all intermediate layers to promote comprehensive point cloud understanding.

To this end, we first analyze the characteristics of the intermediate features. As the layer depth increases, the intermediate features from different layers exhibit a temporally ordered characteristic. Additionally, the total number of intermediate features across layers far exceeds the number of patches in each layer, making traditional attention-based fusion infeasible due to quadratic complexity. Given these two characteristics, we note the Mamba~\cite{mamba} architecture—a recent advancement in state space models designed for long-sequence modeling—as an ideal fit for our needs. Therefore, we propose a novel PEFT method orthogonal to the pre-trained model: Point Mamba Adapter (PMA).

\subsection{Point Mamba Adapter}

\begin{figure*}[htbp]
    \begin{center}
    \includegraphics[width=0.9\linewidth]{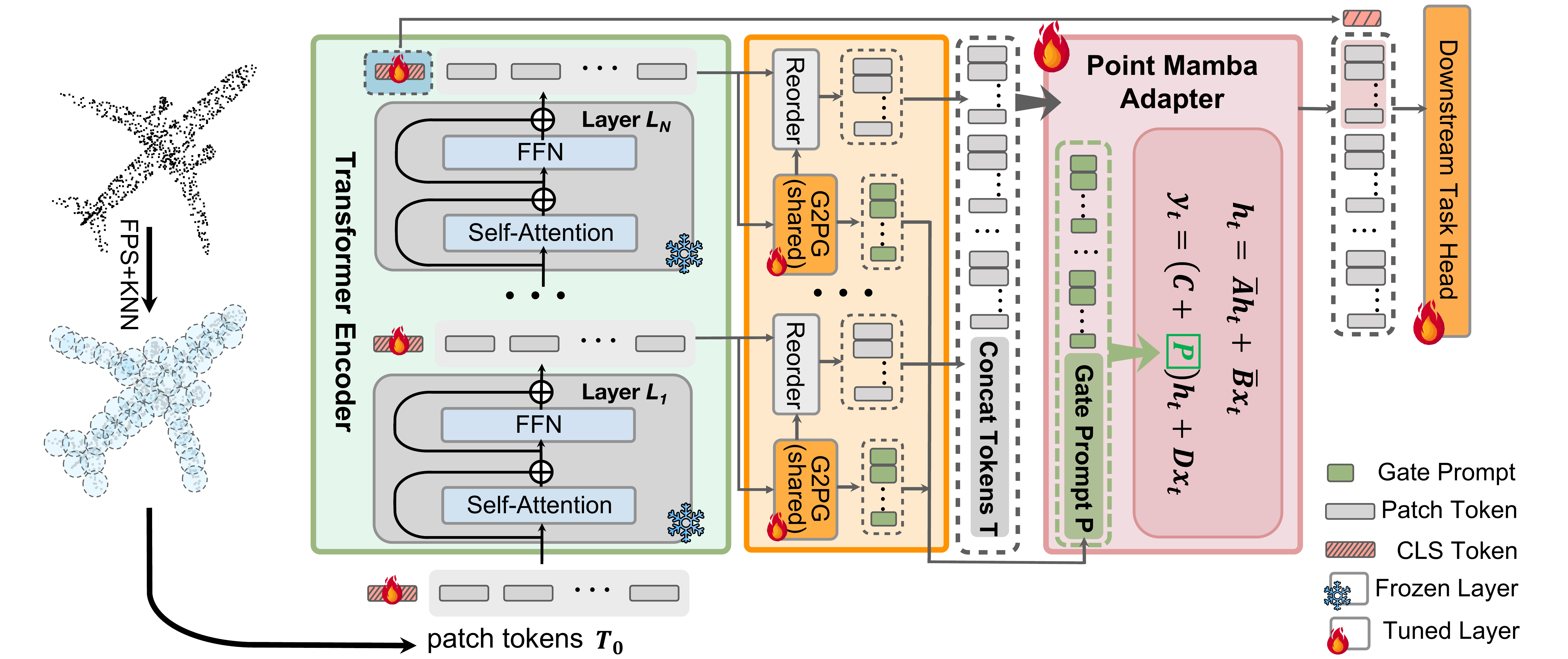}
    \caption{The parameter-efficient fine-tuning pipeline based on our Point Mamba Adapter. It consists of three main components: the pre-trained model, the shared Geometry-constrained Gate Prompt Generator (G2PG), and the Point Mamba Adapter. During fine-tuning, only the CLS token, the G2PG, the Mamba Adapter, and the downstream task head are updated.
    }\label{framework}
    \end{center}
\end{figure*}

The overall framework of our Point Mamba Adapter (PMA) is illustrated in Figure \ref{framework}. It consists of a frozen Transformer backbone network, a shared G2PG module, and a Mamba Adapter that is orthogonal to the backbone. 

\subsection{The Pipeline of Parameter-Efficient Fine-Tuning with Mamba Adapter}

For a given point cloud, the pre-trained model uses Farthest Point Sampling (FPS) and K-Nearest Neighbors (KNN) algorithms to divide it into multiple point patches. A PointNet-based feature extractor is then employed to generate embeddings for these patches, followed by MLPs to produce positional encodings. A \texttt{CLS} $\bm{c_i\in \mathbb{R}^{1\times D}}$is then concatenated with these embeddings and fed into an $L$-layer Transformer backbone. The input tokens $\bm{T_i\in \mathbb{R}^{M\times D}}$ for each layer are obtained by adding the output tokens from the previous layer with their respective positional encodings. Each Transformer layer consists of a Self-Attention module and a Feed-Forward Network (FFN). The forward process of each Transformer layer is defined as:
\begin{gather}
    \label{eq1}
    [\bm c_i; \bm T_i] = l_i([\bm c_{i-1}; \bm T_{i-1}]),\ i=1,2,\cdots,L,
\end{gather}
where $M$ is the number of tokens, and $D$ is the dimensionality of each token. $l_i$ is the $i$-th transformer layer. $L$ is the total number of transformer layers. 

Subsequently, we extract the output features $\bm T_i$ from each layer and feed them into a geometry-constrained gate prompt generator (G2PG) $g$, which is shared across all layers. This module leverages the spatial neighborhood constraints of the point cloud to learn additional geometric prompts $\bm P_i$ for Mamba's output gate $\bm C$. By doing so, it enables Mamba to adjust the current state’s output based on geometric prompts rather than relying solely on preceding inputs, thus achieving a more comprehensive perception during sequence modeling. This process can be formulated as:
\begin{gather}
    \label{eq1}
    \bm P_i = g(\bm T_i),\ i=1,2,\cdots,L,
\end{gather}

Meanwhile, within the G2PG, we also generate an order for tokens at each layer, enabling a clearly defined layer-wise sorting of tokens. Subsequently, we concatenate the tokens from all layers in a chronological sequence to obtain the complete set of tokens $\bm T$ from the pre-trained model. Similarly, we concatenate the geometric prompts from each layer to form a unified prompt $\bm P$ for Mamba’s output gate. These concatenated tokens $\bm T$ and prompts $\bm P$ are then fed into our Mamba Adapter $m$ for comprehensive feature fusion. For the $t$-th token $x_t$ in $\bm T$, its computation can be expressed as:
\begin{gather}
    h_t = \overline{\bm A}h_{t-1} + \overline{\bm B}x_t \\
    y_t = {(\bm C+\textcolor{red}{\bm P})}h_{t} + {\bm D}x_{t}
\end{gather}
where $t$ ranges from 1 to $L \times M$, and $\bm P$ represents the geometric prompt. All 
$y_t$ are concatenated together to form the output $\bm Y$, which represents the fully integrated intermediate variables across all layers.

Finally, the features $\bm F_{pre}$ from the first $N-1$ layers, the final layer's features $\bm F_{last}$, and the final \texttt{[CLS]} $\bm C_N$ token are concatenated and fed into a task-specific head $f$ to generate the predictions $y$ of downstream tasks.
\begin{gather}
    \label{eq2}
    \bm y = f([\bm C_N; \bm F_{last}; \bm F_{pre}]).
\end{gather}

In the parameter-efficient fine-tuning, all parameters of the Transformer backbone $\{l_i\}_{i=1}^L$  are frozen. Only the CLS token $\{C_i\}_{i=1}^L$, the G2PG $g$, Mamba Adapter $m$, and downstream task head $f$ are updated, reducing the demand for storage resources.

\subsection{Geometry-constrained Gate Prompt Generator}

\subsubsection{Motivation}
The motivation for introducing the Geometry-Constrained Gate Prompt Generator (G2PG) arises from several key factors related to the unique characteristics of point cloud data and the challenges of sequence modeling in three-dimensional space.

\textbf{1) Isotropy of 3D Space:} Three-dimensional space exhibits inherent isotropy, meaning there is no intrinsic directionality in the data. This poses a challenge for processing point cloud data and constructing ordered sequences, as traditional Mamba-based methods often rely on predefined directions or sequences. In this context, constructing an ordered sequence becomes particularly difficult since it is not flexible to rely on a fixed spatial order or direction in 3D space.

\textbf{2) Spatial Neighborhood Constraints of Point Cloud Data:} A distinctive feature of point cloud data is the spatial relationship between points, where each point has specific neighborhood relations with surrounding points. The introduction of G2PG allows us to leverage these spatial neighborhood constraints by learning additional geometric prompts, which guide the model to better focus on the spatial relationships within the data, rather than solely depending on the sequence or previous states.

\textbf{3) Improving the Adaptability of the Mamba's Output:} Traditional sequence modeling methods often rely on the preceding state of the sequence to determine the current output. However, by incorporating geometric prompts into the output matrix $\bm C$, we allow the Mamba model to adjust its output based on spatial information. Previous work~\cite{demystify,mambairv2} has already demonstrated that $\bm C$ corresponds to the query in the Transformer mechanism, integrating geometric prompts into this matrix helps the model consider both spatial structure and input data when making predictions. This enhancement enables the model to more effectively "query" unseen or missing regions in the point cloud, thus improving its ability to process spatial dependencies.

\subsubsection{The Pipeline of G2PG}

\begin{figure}[htbp]
\centering
\includegraphics[width=1.0\linewidth]{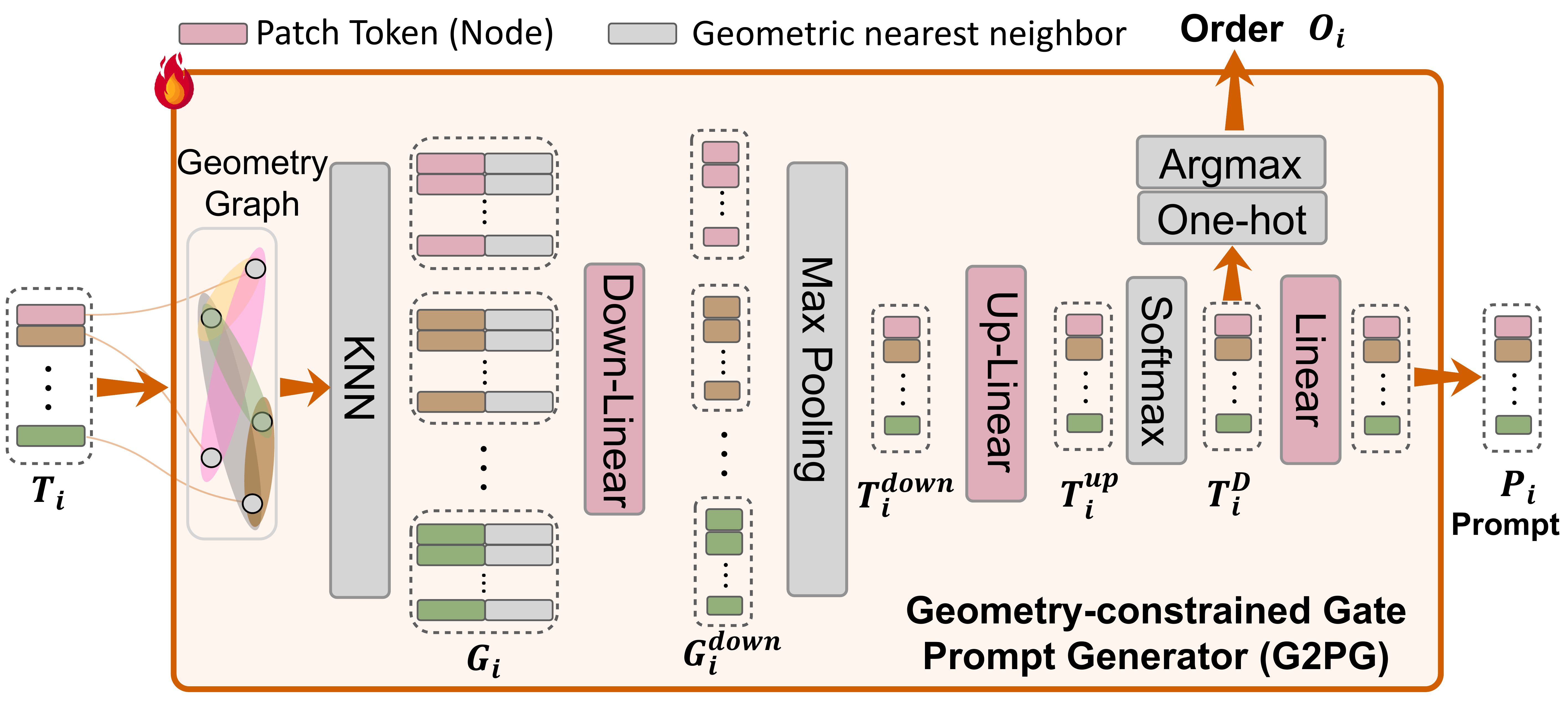} 
\caption{Details of our Geometry-constrained Gate Prompt Generator (G2PG).
}
\label{g2pg}
\end{figure}

Figure \ref{g2pg} illustrates the detail of our Geometry-constrained Gate Prompt Generator, it first constructs a connectivity graph for each token based on the central coordinates in the geometric space using KNN and aggregates the features of each token's neighbors through Down Linear and Max Pooling operations to reinforce the geometric constraints. Subsequently, the features are mapped to the dimensions of the output matrix $\bm C$ using an Up Linear layer. In our experiments, the dimension of $\bm C$ is set to the number of patches, for example, $\bm S = 128$ for the classification task on ScanObjectNN, ensuring that each token can generate its unique identifier. Following this, we apply Softmax to obtain the probability distribution ${\bm T_i^D}\in \mathbb{R}^{m\times S}$. 

In the subsequent process, ${\bm T_i^D}$ serves two primary functions: \textbf{1) Generation of Unique Indices for Geometric Semantic-Aware Sorting:} Initially, ${\bm T_i^D}$ undergoes One-hot encoding followed by an Argmax operation, which assigns a unique index to each token. This process enables geometric semantic-aware sorting, where each token is assigned a specific order based on its position in the geometric space, allowing the model to better capture spatial relationships. \textbf{2) Generation of Geometric Prompts for Enhanced Spatial Perception:} Afterward, ${\bm T_i^D}$ is further mapped to produce geometric prompts that are applied to the output matrix. These geometric prompts introduce implicit geometric constraints into the Mamba Adapter, enabling the model to incorporate spatial structural information during sequence modeling. This enhances the model's perceptual ability, improving its capacity to understand and process the spatial relationships inherent in the point cloud data.

\begin{table*}[h]
  \centering
  \resizebox{\textwidth}{!}{
    \begin{tabular}{lcllllll}
    \toprule
    \multirow{2}[4]{*}{Method} & \multirow{2}[4]{*}{Reference}  & \multicolumn{1}{l}{\multirow{2}[4]{*}{\#TP (M)}}  & \multicolumn{3}{c}{ScanObjectNN} & \multicolumn{2}{c}{ModelNet40} \\
    \cmidrule(lr){4-6}\cmidrule(lr){7-8}\textbf{}        &       &             & OBJ-BG & OBJ-ONLY & PB-T50-RS & w/o Vote & w/ Vote \\
    \midrule
    \multicolumn{8}{c}{\textit{Supervised Learning Only}} \\
    \midrule
    PointNet++ \citep{pointnet++} & NeurIPS 2017    & 1.7   & 82.3  & 84.3  & 77.9  & 90.7  & - \\
    PointMLP \citep{pointmlp} & ICLR 2022     & 12.6  & -     & -     & 85.2  & 94.1  & 94.5 \\
    SFR \citep{sfr}  & ICASSP 2023    &  - &  - &  - & 87.8  & 93.9  & - \\
    P2P-HorNet \citep{p2p} & NeurIPS 2022  & 195.8  & -     & -     & 90.7  & -  & - \\
    X-3D \citep{x3d} & CVPR 2024  & 5.4 & -     & -     & 89.3  & 94.0  & - \\
    \midrule
    \multicolumn{8}{c}{\textit{Self-Supervised Learning (Full fine-tuning)}} \\
    \midrule
    Point-BERT \citep{pointbert} & CVPR 2022   & 22.1  & 87.43 & 88.12 & 83.07 & 92.7  & 93.2 \\
    Point-MAE \citep{pointmae} & ECCV 2022   & 22.1    & 90.02 & 88.29 & 85.18 & 93.2  & 93.8 \\
    Point-M2AE \citep{m2ae} & NeurIPS 2022  & 15.3    & 91.22 & 88.81 & 86.43 & 93.4  & 94.0 \\
    ACT \citep{act}   & ICLR 2023  & 22.1    & 93.29 & 91.91 & 88.21 & 93.2  & 93.7 \\
    I2P-MAE \citep{i2pmae} & CVPR 2023  & 15.3  & 94.15 & 91.57 & 90.11 & 93.7  & 94.1 \\
    Recon \citep{recon} & ICML 2023  & 44.3   & 95.18 & 93.29 & 90.63 & 94.1  & 94.5 \\
    Point-FEMAE \citep{femae} & AAAI 2024  & 27.4  & 95.18 &  93.29 & 90.22 & 94.0  & 94.5 \\
    PointDif \citep{pointdif} & CVPR 2024   & -   & 93.29 & 91.91 & 87.61 & - & -\\
    PointMamba \citep{pointmamba} & NeurIPS 2024  & 12.3  & 94.32 & 92.60 & 89.31 & 93.6 & - \\
    LCM \citep{lcm} & NeurIPS 2024  & 2.7  & 94.51 & 92.75 & 88.87 & 93.6 & 94.2 \\
    MH-PH \citep{mhph} & ECCV 2024  & -  &  97.4 & 96.8 &  93.8 & - & 94.6 \\
    PointGPT-L \citep{pointgpt} (baseline) & NeurIPS 2023  & 360.5  & 97.2 & 96.6 &  93.4 & 94.7  & 94.9 \\
    \midrule
    \multicolumn{8}{c}{\textit{Self-Supervised Learning (Parameter-Efficient fine-tuning)}} \\
    \midrule
     Point-MAE w/ IDPT \citep{idpt} & ICCV 2023  & 2.7   & 95.18 & 93.29 & 90.63 & 94.1  & 94.5 \\
     Recon w/ DAPT \citep{dapt} & CVPR 2024  & 1.1   &  94.32 & 92.43 & 89.38 &  93.5  & 94.1 \\
     PointGPT-L w/ PointGST \citep{pointgst} & arXiv 2024  & 0.6   &  98.97 & 97.59 & 94.83 &   94.8  & 95.3 \\
     \rowcolor{mycolor} \textbf{PointGPT-L w/ PMA (Ours)} & -  & 4.9\textcolor{blue}{($\downarrow$ 99\%)}   &   98.97 \textcolor{blue}{($\uparrow$ 1.77)} & 96.73\textcolor{blue}{($\uparrow$ 0.13)} & 95.18\textcolor{blue}{($\uparrow$ 1.78)} &  94.9\textcolor{blue}{($\uparrow$ 0.2)}  & 95.4\textcolor{blue}{($\uparrow$ 0.5)} \\
    \bottomrule
    \end{tabular}%
  }
  \caption{Classification accuracy on real-scanned (ScanObjectNN) and synthetic (ModelNet40) point clouds. In ScanObjectNN, we report the overall accuracy (\%) on three variants. In ModelNet40, we report the overall accuracy (\%) for both without and with voting. "\#TP(M)" represents the model's trainable parameters.}
  \label{class}%
\end{table*}%

\section{Experiments}

We evaluated the proposed method on tasks including classification, few-shot learning, and part segmentation, using various prominent pre-trained models.

\subsection{Object Point Cloud Classification}

We assess our PMA performance on the ScanObjectNN~\cite{scanobjectnn} and ModelNet40~\cite{modelnet} datasets. ScanObjectNN is a challenging 3D real-world objects dataset that consists of about 15K point cloud samples by 15 categories. These objects are scanned indoor scene data, which are usually cluttered with background and occluded by other objects. We conducted experiments on three variants of ScanObjectNN (OBJ-BG, OBJ-ONLY, and PB-T50-RS).  ModelNet40 includes 12K synthetic point clouds belonging to 40 different categories, with each point cloud being complete and clean, providing a better representation of 3D object shapes.

\begin{table}[t]
  \centering
  \resizebox{\linewidth}{!}{
    \begin{tabular}{lllcccc}
    \toprule
    \multirow{2}[4]{*}{Model} & \multirow{2}[4]{*}{Strategy} & \multirow{2}[4]{*}{\#TP} & \multicolumn{3}{c}{ScanObjectNN} & \multirow{2}[4]{*}{ModelNet40} \\
\cmidrule{4-6}          &       &       & \multicolumn{1}{c}{OBJ-BG} & \multicolumn{1}{c}{OBJ-ONLY} & \multicolumn{1}{c}{PB-T50-RS} & \\
    \midrule
    \multirow{5}[2]{*}{Point-BERT} & FFT   & 22.1  & 87.43 & 88.12 & 83.07 & 92.7\\
          & IDPT \cite{idpt}  & 1.7   & 88.12 & 88.3  & 83.69 & 92.6\\
          & DAPT \cite{dapt}  & 1.1   & 91.05 & 89.67 & 85.43 & 93.1\\
          & PointGST \cite{pointgst}  & 0.6   & 91.39 & 89.67 & \textbf{85.64} & 93.4\\
          \rowcolor{mycolor} & \textbf{PMA}   & 1.1   & \textbf{91.39} & \textbf{91.05} & 85.50 & \textbf{93.7}\\
    \midrule
    \multirow{4}[2]{*}{Point-MAE} & FFT   & 22.1  & 90.02 & 88.29 & 85.18 & 93.2\\
          & IDPT \cite{idpt}  & 1.7   & \textbf{91.22} & 90.02 & 84.94 & 93.3\\
          & DAPT \cite{dapt}  & 1.1   & 90.88 & 90.19 & 85.08 & 93.5\\
          & PointGST \cite{pointgst}  & 0.6   & 91.74 & 90.19 & 85.29 & 93.5\\
          \rowcolor{mycolor} & \textbf{PMA}   & 1.1   & 91.05 & \textbf{90.89} & \textbf{86.43} & \textbf{94.0}\\
    \midrule
    \multirow{4}[2]{*}{PointGPT-L} & FFT   & 360.5 & 97.2  & 96.6  & 93.4 & 94.7\\
          & IDPT \cite{idpt}  & 10    & 98.11 & 96.04 & 92.99 & 94.4\\
          & DAPT \cite{dapt}  & 4.2   & 98.11 & 96.21 & 93.02 & 94.2\\
          & PointGST \cite{pointgst}  & 2.4   & 98.97 & \textbf{97.59} & 94.83 & 94.8\\
          \rowcolor{mycolor} & \textbf{PMA}   & 4.9   & \textbf{98.97} & 96.73 & \textbf{95.18} & \textbf{94.9}\\
    \bottomrule
    \end{tabular}%
  }
  \caption{Compared with other fine-tuning methods. We report the overall accuracy (OA) and
 trainable parameters(\#TP.) across three variants of ScanObjectNN.}
  \label{other}%
\end{table}%

\subsubsection{Compared with state-of-the-art methods} 

We first compare our method with state-of-the-art algorithms, primarily including models based on supervised learning, full fine-tuning of self-supervised models, and parameter-efficient fine-tuning approaches. The state-of-the-art method, PointGPT-L, remains one of the leading self-supervised pre-training models. However, as shown in Table \ref{class}, it has significantly more learnable parameters compared to traditional unsupervised methods (e.g., 360M, which is much larger than X3D's 5.4M). In this paper, we report the performance of our approach based on PointGPT-L, showing a 99\% reduction in learnable parameters compared to the original PointGPT-L full fine-tuning, while also achieving significant improvements in classification accuracy across various datasets. For instance, on the ScanObjectNN PB-T50-RS, our method achieved a 1.78\% increase. 

These results highlight the importance of our parameter-efficient fine-tuning approach, especially given that point cloud models are often deployed on resource-constrained devices such as robots. Compared to fully fine-tuning, parameter-efficient methods can significantly reduce resource requirements. Additionally, by fine-tuning only a subset of parameters, our approach mitigates overfitting during the model adaptation process, thereby enhancing overall performance.

\subsubsection{Compared with other fine-tuning methods} 

We further conduct a detailed comparison of representative fine-tuning strategies used in point cloud tasks across various pre-trained models, including Point-BERT, Point-MAE, and PointGPT-L. These strategies encompass fully fine-tuning (FFT), IDPT, DAPT, PointGST, as well as our proposed PMA method. 

The detailed results are presented in Table \ref{other}, our PMA significantly reduces the number of parameters required for fine-tuning by effectively leveraging intermediate features from pre-trained models. For instance, compared to the fully fine-tuning (FFT) approach, PMA only adjusts a minimal number of trainable parameters while achieving superior performance. On the PointGPT-L pre-trained model, our method requires only 4.9M trainable parameters, which represents a 99\% reduction in parameter count compared to FFT. This makes our PMA highly suitable for deployment on resource-constrained devices, such as embedded systems and robots. Furthermore, compared to existing parameter-efficient fine-tuning methods, PMA uses a similar amount of parameters yet achieves state-of-the-art performance across the majority of models and datasets, demonstrating its efficiency and superior performance.

\subsection{Part Segmentation}
\begin{table}[t]
  \centering
  \resizebox{\linewidth}{!}{
    \begin{tabular}{lclcc}
    \toprule
    Methods & Reference & \#TP (M) &  $\mathrm{mIoU}_{c}$ & $\mathrm{mIoU}_{I}$ \\
    \midrule
    \multicolumn{4}{c}{\textit{Supervised Learning Only}} \\
    \midrule
    PointNet \cite{pointnet} & CVPR 2017 & - & 80.39 & 83.7 \\
    PointNet++  \cite{pointnet++} & NeurIPS 2017 & - & 81.85 & 85.1 \\
    DGCNN \cite{dgcnn} & TOG 2019 & - & 82.33 & 85.2 \\
    \midrule
    \multicolumn{5}{c}{\textit{Self-Supervised Representation Learning (Full fine-tuning)}} \\
    \midrule
    Transformer \cite{attention} & NeurIPS 2017 & 27.09 & 83.42 & 85.1 \\
    MaskPoint \cite{maskpoint} & ECCV 2022 & 27.09 & 84.60 & 86.0 \\
    Point-BERT \cite{pointbert} & CVPR 2022 & 27.09 & 84.11 & 85.6 \\
    Point-MAE \cite{pointmae} & ECCV 2022 & 27.06 & 84.19 & 86.1 \\ 
    ACT \cite{act}  & ICLR 2023 & 27.06 & 84.66 & 86.1 \\
    Recon \cite{recon}  & ICML 2023 & 48.54 & 84.52 & 86.4 \\
    \midrule
    \multicolumn{5}{c}{\textit{Self-Supervised Representation Learning (Efficient fine-tuning)}} \\
    \midrule
    Point-BERT w/ IDPT \cite{idpt} & ICCV 2023 & 5.69 & 83.50  & 85.3  \\
    Point-BERT w/ DAPT \cite{dapt} & CVPR 2024 & 5.65 & 83.83  & 85.5  \\
    Point-BERT w/ PointGST \cite{pointgst} & arXiv 2024 & 5.58 & 83.87  & 85.7  \\
    \rowcolor{mycolor}\textbf{Point-BERT w/ PMA} & Ours & 5.64 &  83.96  & 86.1  \\
    \midrule
    Point-MAE w/ IDPT \cite{idpt} & ICCV 2023 & 5.69 & 83.79  & 85.7  \\
    Point-MAE w/ DAPT \cite{dapt} & CVPR 2024 & 5.65 &  84.01  & 85.7  \\
    Point-MAE w/ PointGST \cite{pointgst} & arXiv 2024 & 5.58 & 83.81  & 85.8  \\
    \rowcolor{mycolor}\textbf{Point-MAE w/ PMA} & Ours & 5.64 &  84.00  & 86.1  \\
    \midrule
    Recon w/ IDPT \cite{idpt} & ICCV 2023 & 5.69 & 83.66  & 85.7  \\
    Recon w/ DAPT \cite{dapt} & CVPR 2024 & 5.65 &  83.87  & 85.7  \\
    Recon w/ PointGST \cite{pointgst} & arXiv 2024 & 5.58 & 83.98  & 85.8  \\
    \rowcolor{mycolor}\textbf{Recon w/ PMA} & Ours & 5.64 &  84.10  & 86.3  \\
    \bottomrule
    \end{tabular}%
    }
  \caption{Part segmentation results on the ShapeNetPart dataset. The mean IoU across all categories, \ie, $\mathrm{mIoU}_{c}$ (\%), and the mean IoU across all instances, \ie, $\mathrm{mIoU}_{I}$ (\%) are reported.}
  \label{part}%
\end{table}%

Part segmentation involves the difficulty of precisely assigning class labels to individual points. In line with DAPT~\cite{dapt}, we adopt Point-BERT, Point-MAE, and Recon as benchmark methods for evaluation on the ShapeNetPart dataset, which contains 16,881 samples distributed across 16 distinct categories. 

Compared to classification, segmentation requires generating independent labels for each point, making it a more finer-grained task. More trainable parameters facilitates fine-grained understanding. For example, as shown in Table \ref{part}, ACT and Point-MAE, having the same number of parameters (\textbf{27M}), result in no improvement for ACT in $\mathrm{mIoU}_{I}$. Recon, which utilizes more parameters (\textbf{49M}), improves $\mathrm{mIoU}_{I}$ by 0.3\%. PEFT imposes a strict constraint on the number of trainable parameters compared to FFT (\textbf{5.6M vs. 27M}), leading to a significant gap (e.g., IDPT and DAPT reduce $\mathrm{mIoU}_{I}$ by 0.4\% compared to FFT). However, our PMA effectively bridges this gap by integrating intermediate features, despite using the same minimal parameters 5.6M.


\subsection{Few-shot Learning}

\begin{table}[htbp]
  \centering
  \resizebox{\linewidth}{!}{
    \begin{tabular}{lcccc}
    \toprule
    & \multicolumn{2}{c}{5-way} & \multicolumn{2}{c}{10-way} \\
\cmidrule{2-5}          & 10-shot & 20-shot & 10-shot & 20-shot \\
    \midrule
    DGCNN-OcCo \cite{wang2021unsupervised}& 90.6±2.8 & 92.5±1.9 & 82.9±1.3 & 86.5±2.2 \\
    Transformer-OcCo \cite{pointbert} & 94.0±3.6 & 95.9±2.3 & 89.4±5.1 & 92.4±4.6 \\
    Point-BERT \cite{pointbert}  & 94.6±3.1 & 96.3±2.7 & 91.0±5.4 & 92.7±5.1 \\
    MaskPoint \cite{maskpoint}  & 95.0±3.7 & 97.2±1.7 & 91.4±4.0 & 93.4±3.5 \\
    Point-MAE \cite{pointmae} & 96.3±2.5 & 97.8±1.8 & 92.6±4.1 & 95.0±3.0 \\
    Point-M2AE \cite{m2ae} & 96.8±1.8 & 98.3±1.4 & 92.3±4.5 & 95.0±3.0 \\
    Recon \cite{recon}  & 97.3±1.9 &  98.9±1.2 & 93.3±3.9 & 93.3±3.9 \\
    PointGPT-L \cite{pointgpt}& 98.0±1.9 &  99.0±1.0 & 94.1±3.3 & 96.1±2.8 \\
    \midrule
    \rowcolor{mycolor}\textbf{Recon w/ PMA}  & 97.4±2.4 &  98.7±1.4 & 94.1±4.0 & 96.2±2.6 \\
    \rowcolor{mycolor}\textbf{PointGPT-L  w/ PMA} & \textbf{98.8±1.2} &  \textbf{99.1±0.9} & \textbf{96.7±2.0} & \textbf{97.5±2.2} \\
    \bottomrule
    \end{tabular}%
  }
  \caption{Few-shot learning on ModelNet40. We report the average classification accuracy (\%) with the standard deviation (\%) of 10 independent experiments.}
  \label{table3}%
\end{table}%

Few-shot learning is a crucial task for evaluating point cloud pre-trained models and fine-tuning methods, particularly when dealing with limited labeled data. Due to the scarcity and high-dimensional complexity of point cloud data, few-shot learning effectively assesses a model's generalization ability and learning efficiency when handling a small number of training samples. We conducted few-shot experiments on ModelNet40, using the n-way, m-shot setting, following previous works \cite{pointmae}. We use Recon and PointGPT-L as our base model. The results for the settings of $n \in { 5, 10 }$ and $m \in {10, 20}$ are presented in Table \ref{table3}.

\subsection{Compare PMA with Other Tuning Strategies} 

\begin{table}[htbp]
  \centering
  \resizebox{\linewidth}{!}{
    \begin{tabular}{lllrr}
    \toprule
    Methods & Reference & Design for & \multicolumn{1}{l}{\#TP (M)} & \multicolumn{1}{l}{PB-T50-RS} \\
    \midrule
    Full Fine-tuning & ECCV 2022 & -     & 22.1  & 85.18 \\
    Linear probing & -     & -     & 0.3   & 75.99 \\
    Adapter \cite{adapter} & ICML 2019 & NLP   & 0.9   & 83.93 \\
    LoRA \cite{lora} & ICLR 2022 & NLP   & 0.9   & 81.74 \\
    BitFit \cite{bitfit} & ACL 2021 & NLP   & 0.3   & 82.62 \\
    VPT \cite{vpt}  & ECCV 2022 & Image & 0.4   & 81.09 \\
    AdaptFormer \cite{adaptformer} & NeurIPS 2022 & Image & 0.9   & 83.45 \\
    BI-AdaptFormer \cite{biadapt} & ICCV 2023 & Image & 0.4   & 83.66 \\
    IDPT \cite{idpt} & ICCV 2023 & Point & 1.7   & 84.94 \\
    DAPT \cite{dapt} & CVPR 2024 & Point & 1.1   & 85.08 \\
    \rowcolor{mycolor} PMA   & Ours  & Point & 1.1   & 86.43 \\
    \bottomrule
    \end{tabular}%
  }
  \caption{Compare PMA with other parameter-efficient fine-tuning.}
  \label{tuning}%
\end{table}%

We further compare the proposed method with the currently mainstream parameter-efficient fine-tuning (PEFT) methods in the language and image domains. Specifically, we directly transfer these methods to fine-tune the pre-trained Point-MAE network without any additional modifications. Table \ref{tuning} presents the experimental results, which clearly show that the methods from the image and language domains exhibit a significant performance gap compared to mainstream point cloud-based PEFT methods. For instance, the best-performing PEFT methods from NLP and 2D vision achieve overall accuracies of 83.93\% and 83.66\%, respectively, which still show a notable performance gap compared to the fully fine-tuned method (85.18\%). This demonstrates that the direct transfer approach is not suitable for the point cloud domain, thus highlighting the necessity of designing task-specific parameter-efficient fine-tuning methods tailored to the characteristics of point cloud data.

\subsection{Ablation Study}

To investigate the architecture design and tuning settings of our proposed strategy, we conducted extensive ablation studies on classification tasks in PB-T50-RS variants of ScanObjectNN \cite{scanobjectnn}.

\subsubsection{The Effect of Each Component}

We conduct experiments to demonstrate the effectiveness of the proposed components in our PMA. Our PMA consists of two main modules: the shared Geometry-Constrained Gate Prompt Generator (G2PG) and the final Mamba Adapter. The G2PG generates two key outputs: one is the gate prompt used to adjust the output matrix $\bm C$ of the Mamba Adapter, and the other is the index order for reordering the hierarchical tokens. We will conduct ablation studies on these key components to assess their individual contributions to the overall performance. 

\begin{table}[htbp]
  \centering
  \resizebox{0.8\linewidth}{!}{
    \begin{tabular}{cccc}
    \toprule
    Mamba Adapter & Gate Prompt & Reorder & PB-T50-RS \\
    \midrule
    \ding{56}     & \ding{56}     & \ding{56}     & 76.72 \\
    \ding{52}     & \ding{56}     & \ding{56}     & 85.08 \\
    \ding{52}     & \ding{52}     & \ding{56}     & 86.02 \\
    \ding{52}     & \ding{52}     & \ding{52}     & 86.43 \\
    \bottomrule
    \end{tabular}%
  }
  \caption{The effect of each component of our PMA. The overall accuracy (\%) on the hardest variant of ScanObjectNN is reported.}
  \label{compo}%
\end{table}%

As shown in Table \ref{compo}, when using a simple Mamba Adapter without any additional reordering, we observe a significant performance improvement compared to using no components at all (76.72\% → 85.08\%), demonstrating the clear advantage of our approach that employs an orthogonal Mamba Adapter to the backbone network. Furthermore, when the gate prompt is introduced, the performance improves even further. This enhancement is attributed to the incorporation of implicit geometric constraints in the Mamba model, which enables it to better perceive previously unseen points. Finally, the introduction of a dynamic reorder mechanism further improves performance, thanks to the task-specific optimization of the ordering process.

\subsubsection{The Effect of Different Ordering Strategies}

Finally, to validate the effectiveness of our end-to-end geometry-constrained dynamic sorting strategy, we further compared it with several common rule-based static sorting algorithms, including sorting along the x, y, and z axes, Hilbert curve \cite{hilbert} sorting, and z-order \cite{zorder} curve sorting. Table \ref{order} reports our experimental results, where we observe that sorting based on the original geometric space (along the x, y, and z axes) yields similar performance. However, since these methods rely entirely on unidirectional scanning, their performance is not the highest. Hilbert curve and Z-ordering also show comparable performance, but they outperform the basic coordinate axis sorting because these spatial sorting methods do not scan each point from a single direction, instead varying the direction for different points. The highest performance is achieved by our end-to-end optimized dynamic sorting strategy, which benefits from the implicit spatial constraints and end-to-end optimization.

\begin{table}[htbp]
  \centering
  \resizebox{0.5\linewidth}{!}{
    \begin{tabular}{lc}
    \toprule
    Order Strategy & \multicolumn{1}{l}{PB-T50-RS} \\
    \midrule
    x-axis & 85.64 \\
    y-axis & 85.49 \\
    z-axis & 85.29 \\
    Hilbert curve & 85.91 \\
    Z-Order curve & 85.95 \\
    Ours  & 86.43 \\
    \bottomrule
    \end{tabular}%
  }
  \caption{The effect of different ordering strategies.}
  \label{order}%
\end{table}%

\section{Conclusion}

We introduced the Point Mamba Adapter (PMA), a novel parameter-efficient fine-tuning solution based on the Mamba architecture. PMA leverages all intermediate features from pre-trained models, maximizing their potential for downstream point cloud tasks. By using state space models to fuse complementary semantics, PMA addresses the challenge of spatial isotropy in 3D space through a Geometry-Constrained Gate Prompt Generator (G2PG). This mechanism optimizes spatial ordering adaptively, enhancing feature integration across layers. Extensive experiments on diverse point cloud datasets demonstrate that PMA significantly boosts point cloud understanding.
\section{Acknowledgment}
This work is supported in part by the National Natural Science Foundation of China, under Grant (62302309,62171248), Shenzhen Science and Technology Program (JCYJ20220818101014030,JCYJ20220818101012025), and the PCNL KEY project (PCL2023AS6-1).
{
    \small
    \bibliographystyle{ieeenat_fullname}
    \bibliography{main}
}


\end{document}